\pdfoutput=1 %

\documentclass[11pt,a4paper]{article}
\usepackage[hyperref]{acl2019}
\usepackage{times}
\usepackage{latexsym}

\usepackage{url}

\usepackage[utf8]{inputenc}
\usepackage[T1]{fontenc} %
\usepackage{multirow}
\usepackage{booktabs}
\usepackage{amsmath}
\usepackage{amsfonts}
\DeclareMathOperator{\score}{score}
\DeclareMathOperator{\similarity}{sim}
\DeclareMathOperator{\levenshtein}{lev}
\DeclareMathOperator{\length}{len}
\DeclareMathOperator{\bleu}{BLEU}
\DeclareMathOperator{\translate}{T}
\DeclareMathOperator{\entropy}{H}
\DeclareMathOperator{\lenpenalty}{LP}
\usepackage{seqsplit}

\aclfinalcopy %

\title{An Effective Approach to Unsupervised Machine Translation}

\author{Mikel Artetxe, Gorka Labaka, Eneko Agirre \\
  IXA NLP Group \\
  University of the Basque Country (UPV/EHU) \\
  \texttt{\{mikel.artetxe, gorka.labaka, e.agirre\}@ehu.eus} \\}

\date{}

\begin{document}
\maketitle
\begin{abstract}
While machine translation has traditionally relied on large amounts of parallel corpora, a recent research line has managed to train both Neural Machine Translation (NMT) and Statistical Machine Translation (SMT) systems using monolingual corpora only. In this paper, we identify and address several deficiencies of existing unsupervised SMT approaches by exploiting subword information, developing a theoretically well founded unsupervised tuning method, and incorporating a joint refinement procedure. Moreover, we use our improved SMT system to initialize a dual NMT model, which is further fine-tuned through on-the-fly back-translation. Together, we obtain large improvements over the previous state-of-the-art in unsupervised machine translation. For instance, we get 22.5 BLEU points in English-to-German WMT 2014, 5.5 points more than the previous best unsupervised system, and 0.5 points more than the (supervised) shared task winner back in 2014.
\end{abstract}

\section{Introduction}
\label{sec:introduction}

The recent advent of neural sequence-to-sequence modeling has resulted in significant progress in the field of machine translation, with large improvements in standard benchmarks \citep{vaswani2017attention,edunov2018understanding} and the first solid claims of human parity in certain settings \citep{hassan2018achieving}. Unfortunately, these systems rely on large amounts of parallel corpora, which are only available for a few combinations of major languages like English, German and French.

Aiming to remove this dependency on parallel data, a recent research line has managed to train unsupervised machine translation systems using monolingual corpora only. The first such systems were based on Neural Machine Translation (NMT), and combined denoising autoencoding and back-translation to train a dual model initialized with cross-lingual embeddings \citep{artetxe2018unmt,lample2018unsupervised}. Nevertheless, these early systems were later superseded by Statistical Machine Translation (SMT) based approaches, which induced an initial phrase-table through cross-lingual embedding mappings, combined it with an n-gram language model, and further improved the system through iterative back-translation \citep{lample2018phrase,artetxe2018usmt}.

In this paper, we develop a more principled approach to unsupervised SMT, addressing several deficiencies of previous systems by incorporating subword information, applying a theoretically well founded unsupervised tuning method, and developing a joint refinement procedure. In addition to that, we use our improved SMT approach to initialize an unsupervised NMT system, which is further improved through on-the-fly back-translation.

Our experiments on WMT 2014/2016 French-English and German-English show the effectiveness of our approach, as our proposed system outperforms the previous state-of-the-art in unsupervised machine translation by 5-7 BLEU points in all these datasets and translation directions. Our system also outperforms the supervised WMT 2014 shared task winner in English-to-German, and is around 2 BLEU points behind it in the rest of translation directions, suggesting that unsupervised machine translation can be a usable alternative in practical settings.

The remaining of this paper is organized as follows. Section \ref{sec:related} first discusses the related work in the topic. Section \ref{sec:smt} then describes our principled unsupervised SMT method, while Section \ref{sec:nmt} discusses our hybridization method with NMT. We then present the experiments done and the results obtained in Section \ref{sec:experiments}, and Section \ref{sec:conclusions} concludes the paper.

\section{Related work}
\label{sec:related}

Early attempts to build machine translation systems with monolingual corpora go back to statistical decipherment \citep{ravi2011deciphering,dou2012large}. These methods see the source language as ciphertext produced by a noisy channel model that first generates the original English text and then probabilistically replaces the words in it. The English generative process is modeled using an n-gram language model, and the channel model parameters are estimated using either expectation maximization or Bayesian inference. This basic approach was later improved by incorporating syntactic knowledge \citep{dou2013dependency} and word embeddings \citep{dou2015unifying}. Nevertheless, these methods were only shown to work in limited settings, being most often evaluated in word-level translation.

More recently, the task got a renewed interest after the concurrent work of \citet{artetxe2018unmt} and \citet{lample2018unsupervised} on unsupervised NMT which, for the first time, obtained promising results in standard machine translation benchmarks using monolingual corpora only. Both methods build upon the recent work on unsupervised cross-lingual embedding mappings, which independently train word embeddings in two languages and learn a linear transformation to map them to a shared space through self-learning \citep{artetxe2017learning,artetxe2018robust} or adversarial training \citep{conneau2018word}. The resulting cross-lingual embeddings are used to initialize a shared encoder for both languages, and the entire system is trained using a combination of denoising autoencoding, back-translation and, in the case of \citet{lample2018unsupervised}, adversarial training. This method was further improved by \citet{yang2018unsupervised}, who use two language-specific encoders sharing only a subset of their parameters, and incorporate a local and a global generative adversarial network. Concurrent to our work, \citet{lample2019cross} report strong results initializing an unsupervised NMT system with a cross-lingual language model.

Following the initial work on unsupervised NMT, it was argued that the modular architecture of phrase-based SMT was more suitable for this problem, and \citet{lample2018phrase} and \citet{artetxe2018usmt} adapted the same principles discussed above to train an unsupervised SMT model, obtaining large improvements over the original unsupervised NMT systems. More concretely, both approaches learn cross-lingual n-gram embeddings from monolingual corpora based on the mapping method discussed earlier, and use them to induce an initial phrase-table that is combined with an n-gram language model and a distortion model. This initial system is then refined through iterative back-translation \citep{sennrich2016improving} which, in the case of \citet{artetxe2018usmt}, is preceded by an unsupervised tuning step. Our work identifies some deficiencies in these previous systems, and proposes a more principled approach to unsupervised SMT that incorporates subword information, uses a theoretically better founded unsupervised tuning method, and applies a joint refinement procedure, outperforming these previous systems by a substantial margin.

Very recently, some authors have tried to combine both SMT and NMT to build hybrid unsupervised machine translation systems. This idea was already explored by \citet{lample2018phrase}, who aided the training of their unsupervised NMT system by combining standard back-translation with synthetic parallel data generated by unsupervised SMT. \citet{marie2018unsupervised} go further and use synthetic parallel data from unsupervised SMT to train a conventional NMT system from scratch. The resulting NMT model is then used to augment the synthetic parallel corpus through back-translation, and a new NMT model is trained on top of it from scratch, repeating the process iteratively. \citet{ren2019unsupervised} follow a similar approach, but use SMT as posterior regularization at each iteration. As shown later in our experiments, our proposed NMT hybridization obtains substantially larger absolute gains than all these previous approaches, even if our initial SMT system is stronger and thus more challenging to improve upon.

\section{Principled unsupervised SMT}
\label{sec:smt}

Phrase-based SMT is formulated as a log-linear combination of several statistical models: a translation model, a language model, a reordering model and a word/phrase penalty. As such, building an unsupervised SMT system requires learning these different components from monolingual corpora. As it turns out, this is straightforward for most of them: the language model is learned from monolingual corpora by definition; the word and phrase penalties are parameterless; and one can drop the standard lexical reordering model at a small cost and do with the distortion model alone, which is also parameterless. This way, the main challenge left is learning the translation model, that is, building the phrase-table.

Our proposed method starts by building an initial phrase-table through cross-lingual embedding mappings (Section \ref{subsec:initial}). This initial phrase-table is then extended by incorporating subword information, addressing one of the main limitations of previous unsupervised SMT systems (Section \ref{subsec:subword}). Having done that, we adjust the weights of the underlying log-linear model through a novel unsupervised tuning procedure (Section \ref{subsec:tuning}). Finally, we further improve the system by jointly refining two models in opposite directions (Section \ref{subsec:refinement}).

\subsection{Initial phrase-table} \label{subsec:initial}

So as to build our initial phrase-table, we follow \citet{artetxe2018usmt} and learn n-gram embeddings for each language independently, map them to a shared space through self-learning, and use the resulting cross-lingual embeddings to extract and score phrase pairs.

More concretely, we train our n-gram embeddings using \textit{phrase2vec}\footnote{\url{https://github.com/artetxem/phrase2vec}}, a simple extension of skip-gram that applies the standard negative sampling loss of \citet{mikolov2013distributed} to bigram-context and trigram-context pairs in addition to the usual word-context pairs.\footnote{So as to keep the model size within a reasonable limit, we restrict the vocabulary to the most frequent 200,000 unigrams, 400,000 bigrams and 400,000 trigrams.} Having done that, we map the embeddings to a cross-lingual space using VecMap\footnote{\url{https://github.com/artetxem/vecmap}} with \textit{identical} initialization \citep{artetxe2018robust}, which builds an initial solution by aligning identical words and iteratively improves it through self-learning. Finally, we extract translation candidates by taking the 100 nearest-neighbors of each source phrase, and score them by applying the softmax function over their cosine similarities:
\[ \phi ( \bar{f} | \bar{e} ) = \frac{ \exp \left( \cos( \bar{e}, \bar{f} ) / \tau \right)}{\sum_{\bar{f'}} \exp \left(  \cos ( \bar{e}, \bar{f'} ) / \tau \right)} \]
where the temperature $\tau$ is estimated using maximum likelihood estimation over a dictionary induced in the reverse direction. In addition to the phrase translation probabilities in both directions, the forward and reverse lexical weightings are also estimated by aligning each word in the target phrase with the one in the source phrase most likely generating it, and taking the product of their respective translation probabilities. The reader is referred to \citet{artetxe2018usmt} for more details.

\subsection{Adding subword information} \label{subsec:subword}

An inherent limitation of existing unsupervised SMT systems is that words are taken as atomic units, making it impossible to exploit character-level information. This is reflected in the known difficulty of these models to translate named entities, as it is very challenging to discriminate among related proper nouns based on distributional information alone, yielding to translation errors like \textit{``Sunday Telegraph''} $\rightarrow$ \textit{``The Times of London''} \citep{artetxe2018usmt}.

So as to overcome this issue, we propose to incorporate subword information once the initial alignment is done at the word/phrase level. For that purpose, we add two additional weights to the initial phrase-table that are analogous to the lexical weightings, but use a character-level similarity function instead of word translation probabilities:
\[ \score ( \bar{f} | \bar{e} ) = \prod_i \max \left( \epsilon, \max_j \similarity (\bar{f}_i, \bar{e}_j) \right) \]
where $\epsilon=0.3$ guarantees a minimum similarity score, as we want to favor translation candidates that are similar at the character level without excessively penalizing those that are not. In our case, we use a simple similarity function that normalizes the Levenshtein distance $\levenshtein (\cdot)$  \citep{levenshtein1966binary} by the length of the words $\length (\cdot)$:
\[ \similarity (f, e) = 1 - \frac{\levenshtein (f, e)}{\max ( \length (f), \length (e))} \]
We leave the exploration of more elaborated similarity functions and, in particular, learnable metrics \citep{mccallum2005conditional}, for future work.

\subsection{Unsupervised tuning} \label{subsec:tuning}

Having trained the underlying statistical models independently, SMT tuning aims to adjust the weights of their resulting log-linear combination to optimize some evaluation metric like BLEU in a parallel validation corpus, which is typically done through Minimum Error Rate Training or MERT \citep{och2003MERT}. Needless to say, this cannot be done in strictly unsupervised settings, but we argue that it would still be desirable to optimize some unsupervised criterion that is expected to correlate well with test performance. Unfortunately, neither of the existing unsupervised SMT systems do so: \citet{artetxe2018usmt} use a heuristic that builds two initial models in opposite directions, uses one of them to generates a synthetic parallel corpus through back-translation \citep{sennrich2016improving}, and applies MERT to tune the model in the reverse direction, iterating until convergence, whereas \citet{lample2018phrase} do not perform any tuning at all. In what follows, we propose a more principled approach to tuning that defines an unsupervised criterion and an optimization procedure that is guaranteed to converge to a local optimum of it.

Inspired by the previous work on CycleGANs \citep{zhu2017unpaired} and dual learning \citep{he2016dual}, our method takes two initial models in opposite directions, and defines an \textbf{unsupervised optimization objective} that combines a cyclic consistency loss and a language model loss over the two monolingual corpora $E$ and $F$:

\[L = L_{cycle}(E) + L_{cycle}(F) + L_{lm}(E) + L_{lm}(F)\]

The cyclic consistency loss captures the intuition that the translation of a translation should be close to the original text. So as to quantify this, we take a monolingual corpus in the source language, translate it to the target language and back to the source language, and compute its BLEU score taking the original text as reference:
\[L_{cycle}(E) = 1 - \bleu (\translate_{F \rightarrow E} ( \translate_{E \rightarrow F} (E)), E) \]

At the same time, the language model loss captures the intuition that machine translation should produce fluent text in the target language. For that purpose, we estimate the per-word entropy in the target language corpus using an n-gram language model, and penalize higher per-word entropies in machine translated text as follows:\footnote{We initially tried to directly minimize the entropy of the generated text, but this worked poorly in our preliminary experiments on English-Spanish (note that we used this language pair exclusively for development to be faithful to our unsupervised scenario at test time). More concretely, the behavior of the optimization algorithm was very unstable, as it tended to excessively focus on either the cyclic consistency loss or the language model loss at the cost of the other, and we found it very difficult to find the right balance between the two factors.}
\[L_{lm}(E) = \lenpenalty \cdot \max (0, \entropy (F) - \entropy (\translate_{E \rightarrow F} (E)) ) ^2 \]
where the length penalty $\lenpenalty = \lenpenalty (E) \cdot \lenpenalty (F)$ penalizes excessively long translations:\footnote{Without this penalization, the system tended to produce unnecessary tokens (e.g. quotes) that looked natural in their context, which served to minimize the per-word perplexity of the output. Minimizing the overall perplexity instead of the per-word perplexity did not solve the problem, as the opposite phenomenon arose (i.e. the system tended to produce excessively short translations).}
\[ \lenpenalty (E) = \max \left( 1, \frac{\length (\translate_{F \rightarrow E} ( \translate_{E \rightarrow F} (E)))}{\length (E)} \right) \]

So as to minimize the combined loss function, we \textbf{adapt MERT to jointly optimize} the parameters of the two models. In its basic form, MERT approximates the search space for each source sentence through an n-best list, and performs a form of coordinate descent by computing the optimal value for each parameter through an efficient line search method and greedily taking the step that leads to the largest gain. The process is repeated iteratively until convergence, augmenting the n-best list with the updated parameters at each iteration so as to obtain a better approximation of the full search space.
Given that our optimization objective combines two translation systems $\translate_{F \rightarrow E} ( \translate_{E \rightarrow F} (E))$, this would require generating an n-best list for $\translate_{E \rightarrow F} (E)$ first and, for each entry on it, generating a new n-best list with $\translate_{F \rightarrow E}$, yielding a combined n-best list with $N^2$ entries. So as to make it more efficient, we propose an alternating optimization approach where we fix the parameters of one model and optimize the other with standard MERT. Thanks to this, we do not need to expand the search space of the fixed model, so we can do with an n-best list of $N$ entries alone. Having done that, we fix the parameters of the opposite model and optimize the other, iterating until convergence.

\subsection{Joint refinement} \label{subsec:refinement}

Constrained by the lack of parallel corpora, the procedure described so far makes important simplifications that could compromise its potential performance: its phrase-table is somewhat unnatural (e.g. the translation probabilities are estimated from cross-lingual embeddings rather than actual frequency counts) and it lacks a lexical reordering model altogether. So as to overcome this issue, existing unsupervised SMT methods generate a synthetic parallel corpus through back-translation and use it to train a standard SMT system from scratch, iterating until convergence.

An obvious drawback of this approach is that the back-translated side will contain ungrammatical n-grams and other artifacts that will end up in the induced phrase-table. One could argue that this should be innocuous as long as the ungrammatical n-grams are in the source side, as they should never occur in real text and their corresponding entries in the phrase-table should therefore not be used. However, ungrammatical source phrases do ultimately affect the estimation of the backward translation probabilities, including those of grammatical phrases.\footnote{For instance, let's say that the target phrase \textit{``dos gatos''} has been aligned 10 times with \textit{``two cats''} and 90 times with \textit{``two cat''}. While the ungrammatical phrase-table entry \textit{two cat- dos gatos} should never be picked, the backward probability estimation of \textit{two cats - dos gatos} is still affected by it (it would be 0.1 instead of 1.0 in this example).} We argue that, ultimately, the backward probability estimations can only be meaningful when all source phrases are grammatical (so the probabilities of all plausible translations sum to one) and, similarly, the forward probability estimations can only be meaningful when all target phrases are grammatical.

Following the above observation, we propose an alternative approach that jointly refines both translation directions. More concretely, we use the initial systems to build two synthetic corpora in opposite directions.\footnote{For efficiency purposes, we restrict the size of each synthetic parallel corpus to 10 million sentence pairs.} Having done that, we independently extract phrase pairs from each synthetic corpus, and build a phrase-table by taking their intersection. The forward probabilities are estimated in the parallel corpus with the synthetic source side, while the backward probabilities are estimated in the one with the synthetic target side. This does not only guarantee that the probability estimates are meaningful as discussed previously, but it also discards the ungrammatical phrases altogether, as both the source and the target n-grams must have occurred in the original monolingual texts to be present in the resulting phrase-table. This phrase-table is then combined with a lexical reordering model learned on the synthetic parallel corpus in the reverse direction, and we apply the unsupervised tuning method described in Section \ref{subsec:tuning} to adjust the weights of the resulting system. We repeat this process for a total of 3 iterations.\footnote{For the last iteration, we do not perform any tuning and use default Moses weights instead, which we found to be more robust during development. Note, however, that using unsupervised tuning during the previous steps was still strongly beneficial.}

\begin{table*}[t]
\begin{center}
\begin{small}
  \begin{tabular}{clccccccc}
    \toprule
    & & \multicolumn{4}{c}{WMT-14} & & \multicolumn{2}{c}{WMT-16} \\
    \cmidrule{3-6} \cmidrule{8-9}
    & & \multicolumn{1}{c}{fr-en} & \multicolumn{1}{c}{en-fr} & \multicolumn{1}{c}{de-en} & \multicolumn{1}{c}{en-de} & & \multicolumn{1}{c}{de-en} & \multicolumn{1}{c}{en-de} \\
    \midrule
    \multirow{4}{*}{NMT}
    & \citet{artetxe2018unmt} & 15.6 & 15.1 & 10.2 & 6.6 & & - & - \\
    & \citet{lample2018unsupervised} & 14.3 & 15.1 & - & - & & 13.3 & 9.6 \\
    & \citet{yang2018unsupervised} & 15.6 & 17.0 & - & - & & 14.6 & 10.9 \\
    & \citet{lample2018phrase} & \underline{24.2} & \underline{25.1} & - & - & & \underline{21.0} & \underline{17.2} \\
    \midrule
    \multirow{5}{*}{SMT}
    & \citet{artetxe2018usmt} & 25.9 & 26.2 & 17.4 & 14.1 & & 23.1 & 18.2 \\
    & \citet{lample2018phrase} & 27.2 & 28.1 & - & - & & 22.9 & 17.9 \\
    & \citet{marie2018unsupervised}$^*$ & - & - & - & - & & 20.2 & 15.5 \\
    & Proposed system & \underline{28.4} & \underline{30.1} & \underline{20.1} & \underline{15.8} & & \underline{25.4} & \underline{19.7} \\
    & \quad \textit{detok. SacreBLEU}$^*$ & 27.9 & 27.8 & 19.7 & 14.7 & & 24.8 & 19.4 \\
    \midrule
    \multirow{5}{*}{\shortstack{SMT\\+\\NMT}}
    & \citet{lample2018phrase} & 27.7 & 27.6 & - & - & & 25.2 & 20.2 \\
    & \citet{marie2018unsupervised}$^*$ & - & - & - & - & & 26.7 & 20.0 \\
    & \citet{ren2019unsupervised} & 28.9 & 29.5 & 20.4 & 17.0 & & 26.3 & 21.7 \\
    & Proposed system & \bf \underline{33.5} & \bf \underline{36.2} & \bf \underline{27.0} & \bf \underline{22.5} & & \bf \underline{34.4} & \bf \underline{26.9} \\
    & \quad \textit{detok. SacreBLEU}$^*$ & 33.2 & 33.6 & 26.4 & 21.2 & & 33.8 & 26.4 \\
    \bottomrule
  \end{tabular}
\end{small}
\end{center}
\caption{Results of the proposed method in comparison to previous work (BLEU). Overall best results are in bold, the best ones in each group are underlined. \\
$^*$Detokenized BLEU equivalent to the official \texttt{mteval-v13a.pl} script. The rest use tokenized BLEU with \texttt{multi-bleu.perl} (or similar).}
\label{tab:results_main}
\end{table*}

\section{NMT hybridization}
\label{sec:nmt}

While the rigid and modular design of SMT provides a very suitable framework for unsupervised machine translation, NMT has shown to be a fairly superior paradigm in supervised settings, outperforming SMT by a large margin in standard benchmarks. As such, the choice of SMT over NMT also imposes a hard ceiling on the potential performance of these approaches, as unsupervised SMT systems inherit the very same limitations of their supervised counterparts (e.g. the locality and sparsity problems). For that reason, we argue that SMT provides a more appropriate architecture to find an initial alignment between the languages, but NMT is ultimately a better architecture to model the translation process.

Following this observation, we propose a hybrid approach that uses unsupervised SMT to warm up a dual NMT model trained through iterative back-translation. More concretely, we first train two SMT systems in opposite directions as described in Section \ref{sec:smt}, and use them to assist the training of another two NMT systems in opposite directions. These NMT systems are trained following an iterative process where, at each iteration, we alternately update the model in each direction by performing a single pass over a synthetic parallel corpus built through back-translation \citep{sennrich2016improving}.\footnote{Note that we do not train a new model from scratch each time, but continue training the model from the previous iteration.} In the first iteration, the synthetic parallel corpus is entirely generated by the SMT system in the opposite direction but, as training progresses and the NMT models get better, we progressively switch to a synthetic parallel corpus generated by the reverse NMT model. More concretely, iteration $t$ uses $N_{smt} = N \cdot \max (0, 1 - t / a)$ synthetic parallel sentences from the reverse SMT system, where the parameter $a$ controls the number of transition iterations from SMT to NMT back-translation. The remaining $N - N_{smt}$ sentences are generated by the reverse NMT model. Inspired by \citet{edunov2018understanding}, we use greedy decoding for half of them, which produces more fluent and predictable translations, and random sampling for the other half, which produces more varied translations.
In our experiments, we use $N=1,000,000$ and $a=30$, and perform a total of 60 such iterations. At test time, we use beam search decoding with an ensemble of all checkpoints from every 10 iterations.

\section{Experiments and results}
\label{sec:experiments}

In order to make our experiments comparable to previous work, we use the French-English and German-English datasets from the WMT 2014 shared task. More concretely, our training data consists of the concatenation of all News Crawl monolingual corpora from 2007 to 2013, which make a total of 749 million tokens in French, 1,606 millions in German, and 2,109 millions in English, from which we take a random subset of 2,000 sentences for tuning (Section \ref{subsec:tuning}). Preprocessing is done using standard Moses tools, and involves punctuation normalization, tokenization with aggressive hyphen splitting, and truecasing.

Our SMT implementation is based on Moses\footnote{\url{http://www.statmt.org/moses/}}, and we use the KenLM \citep{heafield2013scalable} tool included in it to estimate our 5-gram language model with modified Kneser-Ney smoothing. Our unsupervised tuning implementation is based on Z-MERT \citep{zaidan2009zmert}, and we use FastAlign \citep{dyer2013simple} for word alignment within the joint refinement procedure. Finally, we use the big transformer implementation from fairseq\footnote{\url{https://github.com/pytorch/fairseq}} for our NMT system, training with a total batch size of 20,000 tokens across 8 GPUs with the exact same hyperparameters as \citet{ott2018scaling}.

We use newstest2014 as our test set for French-English, and both newstest2014 and newstest2016 (from WMT 2016\footnote{Note that it is only the test set that is from WMT 2016. All the training data comes from WMT 2014 News Crawl, so it is likely that our results could be further improved by using the more extensive monolingual corpora from WMT 2016.}) for German-English. Following common practice, we report tokenized BLEU scores as computed by the \texttt{multi-bleu.perl} script included in Moses. In addition to that, we also report detokenized BLEU scores as computed by SacreBLEU\footnote{SacreBLEU signature: \seqsplit{BLEU+case.mixed+lang.LANG +numrefs.1+smooth.exp+test.TEST+tok.13a+version.1.2.11}, with LANG $\in$ \{fr-en, en-fr, de-en, en-de\} and TEST $\in$ \{wmt14/full, wmt16\}} \citep{post2018call}, which is equivalent to the official \texttt{mteval-v13a.pl} script.

We next present the results of our proposed system in comparison to previous work in Section \ref{subsec:results_main}. Section \ref{subsec:results_supervised} then compares the obtained results to those of different supervised systems. Finally, Section \ref{subsec:results_examples} presents some translation examples from our system.

\begin{table*}[t]
\begin{center}
\begin{small}
  \begin{tabular}{llllcll}
    \toprule
    & & \multicolumn{2}{c}{WMT-14} & & \multicolumn{2}{c}{WMT-16} \\
    \cmidrule{3-4} \cmidrule{6-7}
    & & \multicolumn{1}{l}{fr-en} & \multicolumn{1}{l}{en-fr} & & \multicolumn{1}{l}{de-en} & \multicolumn{1}{l}{en-de} \\
    \midrule
    \multirow{2}{*}{\citet{lample2018phrase}}
    & Initial SMT & 27.2 & 28.1 & & 22.9 & 17.9 \\
    & + NMT hybrid & 27.7 \scriptsize{(+0.5)} & 27.6 \scriptsize{(-0.5)} & & 25.2 \scriptsize{(+2.3)} & 20.2 \scriptsize{(+2.3)} \\
    \midrule
    \multirow{2}{*}{\citet{marie2018unsupervised}}
    & Initial SMT & - & - & & 20.2 & 15.5 \\
    & + NMT hybrid & - & - & & 26.7 \scriptsize{(+6.5)} & 20.0 \scriptsize{(+4.5)} \\
    \midrule
    \multirow{2}{*}{Proposed system}
    & Initial SMT & 28.4 & 30.1 & & 25.4 & 19.7 \\
    & + NMT hybrid & \bf 33.5 \scriptsize{(+5.1)} & \bf 36.2 \scriptsize{(+6.1)} & & \bf 34.4 \scriptsize{(+9.0)} & \bf 26.9 \scriptsize{(+7.2)} \\
    \bottomrule
  \end{tabular}
\end{small}
\end{center}
\caption{NMT hybridization results for different unsupervised machine translation systems (BLEU).}
\label{tab:results_hybrid}
\end{table*}

\begin{table*}[t]
\begin{center}
\begin{small}
  \begin{tabular}{llcccl}
    \toprule
    & & \multicolumn{4}{c}{WMT-14} \\
    \cmidrule{3-6}
    & & fr-en & en-fr & de-en & en-de \\
    \midrule
    \multirow{2}{*}{Unsupervised}
    & Proposed system & 33.5 & 36.2 & 27.0 & 22.5 \\
    & \quad \textit{detok. SacreBLEU}$^*$ & 33.2 & 33.6 & 26.4 & 21.2 \\
    \midrule
    \multirow{3}{*}{Supervised}
    & WMT best$^*$ & 35.0 & 35.8 & 29.0 & 20.6$^\dagger$ \\
    & \citet{vaswani2017attention} & - & 41.0 & - & 28.4 \\
    & \citet{edunov2018understanding} & - & 45.6 & - & 35.0 \\
    \bottomrule
  \end{tabular}
\end{small}
\end{center}
\caption{Results of the proposed method in comparison to different supervised systems (BLEU). \\
$^*$Detokenized BLEU equivalent to the official \texttt{mteval-v13a.pl} script. The rest use tokenized BLEU with \texttt{multi-bleu.perl} (or similar). \\
$^\dagger$Results in the original test set from WMT 2014, which slightly differs from the full test set used in all subsequent work. Our proposed system obtains 22.4 BLEU points (21.1 detokenized) in that same subset.}
\label{tab:results_supervised}
\end{table*}

\begin{table*}[t]
\begin{small}
\begin{center}
\addtolength{\tabcolsep}{-0.5pt}
  \begin{tabular}{p{3.82cm}p{3.46cm}p{3.48cm}p{3.67cm}}
    \toprule
    \bf Source & \bf Reference & \bf \citet{artetxe2018usmt} & \bf Proposed system \\
    \midrule
    D'autres révélations ont fait état de documents divulgués par Snowden selon lesquels la NSA avait intercepté des données et des communications émanant du téléphone portable de la chancelière allemande Angela Merkel et de ceux de 34 autres chefs d'État.
    & Other revelations cited documents leaked by Snowden that the NSA monitored German Chancellor Angela Merkel's cellphone and those of up to 34 other world leaders.
    & Other disclosures have reported documents disclosed by Snowden suggested the NSA had intercepted communications and data from the mobile phone of German Chancellor Angela Merkel and those of 32 other heads of state.
    & Other revelations have pointed to documents disclosed by Snowden that the NSA had intercepted data and communications emanating from German Chancellor Angela Merkel's mobile phone and those of 34 other heads of state.
    \\
    \midrule
	La NHTSA n'a pas pu examiner la lettre d'information aux propriétaires en raison de l'arrêt de 16 jours des activités gouvernementales, ce qui a ralenti la croissance des ventes de véhicules en octobre.
	& NHTSA could not review the owner notification letter due to the 16-day government shutdown, which tempered auto sales growth in October.
	& The NHTSA could not consider the letter of information to owners because of halting 16-day government activities, which slowed the growth in vehicle sales in October.
	& NHTSA said it could not examine the letter of information to owners because of the 16-day halt in government operations, which slowed vehicle sales growth in October.
	\\
    \midrule
    Le M23 est né d'une mutinerie, en avril 2012, d'anciens rebelles, essentiellement tutsi, intégrés dans l'armée en 2009 après un accord de paix.
    & The M23 was born of an April 2012 mutiny by former rebels, principally Tutsis who were integrated into the army in 2009 following a peace agreement.
    & M23 began as a mutiny in April 2012, former rebels, mainly Tutsi integrated into the national army in 2009 after a peace deal.
    & The M23 was born into a mutiny in April 2012, of former rebels, mostly Tutsi, embedded in the army in 2009 after a peace deal.
    \\
    \midrule
	Tunks a déclaré au Sunday Telegraph de Sydney que toute la famille était <<extrêmement préoccupée>> du bien-être de sa fille et voulait qu'elle rentre en Australie.
	& Tunks told Sydney's Sunday Telegraph the whole family was ``extremely concerned'' about his daughter's welfare and wanted her back in Australia.
	& Tunks told The Times of London from Sydney that the whole family was ``extremely concerned'' of the welfare of her daughter and wanted it to go in Australia.
	& Tunks told the Sunday Telegraph in Sydney that the whole family was ``extremely concerned'' about her daughter's well-being and wanted her to go into Australia.
	\\
    \bottomrule
  \end{tabular}
\end{center}
\end{small}
\caption{Randomly chosen translation examples from French$\rightarrow$English newstest2014 in comparison of those reported by \citet{artetxe2018usmt}.} \label{tab:examples}
\end{table*}

\subsection{Main results} \label{subsec:results_main}

Table \ref{tab:results_main} reports the results of the proposed system in comparison to previous work. As it can be seen, our full system obtains the best published results in all cases, outperforming the previous state-of-the-art by 5-7 BLEU points in all datasets and translation directions.

A substantial part of this improvement comes from our more principled unsupervised SMT approach, which outperforms all previous SMT-based systems by around 2 BLEU points. Nevertheless, it is the NMT hybridization that brings the largest gains, improving the results of this initial SMT systems by 5-9 BLEU points. As shown in Table \ref{tab:results_hybrid}, our absolute gains are considerably larger than those of previous hybridization methods, even if our initial SMT system is substantially better and thus more difficult to improve upon. This way, our initial SMT system is about 4-5 BLEU points above that of \citet{marie2018unsupervised}, yet our absolute gain on top of it is around 2.5 BLEU points higher. When compared to \citet{lample2018phrase}, we obtain an absolute gain of 5-6 BLEU points in both French-English directions while they do not get any clear improvement, and we obtain an improvement of 7-9 BLEU points in both German-English directions, in contrast with the 2.3 BLEU points they obtain.

More generally, it is interesting that pure SMT systems perform better than pure NMT systems, yet the best results are obtained by initializing an NMT system with an SMT system. This suggests that the rigid and modular architecture of SMT might be more suitable to find an initial alignment between the languages, but the final system should be ultimately based on NMT for optimal results.

\subsection{Comparison with supervised systems} \label{subsec:results_supervised}

So as to put our results into perspective, Table \ref{tab:results_supervised} reports the results of different supervised systems in the same WMT 2014 test set. More concretely, we include the best results from the shared task itself, which reflect the state-of-the-art in machine translation back in 2014; those of \citet{vaswani2017attention}, who introduced the now predominant transformer architecture; and those of \citet{edunov2018understanding}, who apply back-translation at a large scale and, to the best of our knowledge, hold the current best results in the test set.

As it can be seen, our unsupervised system outperforms the WMT 2014 shared task winner in English-to-German, and is around 2 BLEU points behind it in the other translation directions.
This shows that unsupervised machine translation is already competitive with the state-of-the-art in supervised machine translation in 2014. While the field of machine translation has undergone great progress in the last 5 years, and the gap between our unsupervised system and the current state-of-the-art in supervised machine translation is still large as reflected by the other results, this suggests that unsupervised machine translation can be a usable alternative in practical settings.

\subsection{Qualitative results} \label{subsec:results_examples}

Table \ref{tab:examples} shows some translation examples from our proposed system in comparison to those reported by \citet{artetxe2018usmt}. We choose the exact same sentences reported by \citet{artetxe2018usmt}, which were randomly taken from newstest2014, so they should be representative of the general behavior of both systems.

While not perfect, our proposed system produces generally fluent translations that accurately capture the meaning of the original text. Just in line with our quantitative results, this suggests that unsupervised machine translation can be a usable alternative in practical settings.

Compared to \citet{artetxe2018usmt}, our translations are generally more fluent, which is not surprising given that they are produced by an NMT system rather than an SMT system. In addition to that, the system of \citet{artetxe2018usmt} has some adequacy issues when translating named entities and numerals (e.g.  \textit{34} $\rightarrow$ \textit{32}, \textit{Sunday Telegraph} $\rightarrow$ \textit{The Times of London}), which we do not observe for our proposed system in these examples.

\section{Conclusions and future work}
\label{sec:conclusions}

In this paper, we identify several deficiencies in previous unsupervised SMT systems, and propose a more principled approach that addresses them by incorporating subword information, using a theoretically well founded unsupervised tuning method, and developing a joint refinement procedure. In addition to that, we use our improved SMT approach to initialize a dual NMT model that is further improved through on-the-fly back-translation. Our experiments show the effectiveness of our approach, as we improve the previous state-of-the-art in unsupervised machine translation by 5-7 BLEU points in French-English and German-English WMT 2014 and 2016. Our code is available as an open source project at \url{https://github.com/artetxem/monoses}.

In the future, we would like to explore learnable similarity functions like the one proposed by \citep{mccallum2005conditional} to compute the character-level scores in our initial phrase-table. In addition to that, we would like to incorporate a language modeling loss during NMT training similar to \citet{he2016dual}. Finally, we would like to adapt our approach to more relaxed scenarios with multiple languages and/or small parallel corpora.

\section*{Acknowledgments}

This research was partially supported by the Spanish MINECO (UnsupNMT TIN2017‐91692‐EXP and DOMINO PGC2018-102041-B-I00, cofunded by EU FEDER), the BigKnowledge project (BBVA foundation grant 2018), the UPV/EHU (excellence research group), and the NVIDIA GPU grant program. Mikel Artetxe was supported by a doctoral grant from the Spanish MECD.

\bibliography{acl2019}

\begin{thebibliography}{30}
\expandafter\ifx\csname natexlab\endcsname\relax\def\natexlab#1{#1}\fi

\bibitem[{Artetxe et~al.(2017)Artetxe, Labaka, and
  Agirre}]{artetxe2017learning}
Mikel Artetxe, Gorka Labaka, and Eneko Agirre. 2017.
\newblock \href {http://aclweb.org/anthology/P17-1042} {Learning bilingual word
  embeddings with (almost) no bilingual data}.
\newblock In \emph{Proceedings of the 55th Annual Meeting of the Association
  for Computational Linguistics (Volume 1: Long Papers)}, pages 451--462,
  Vancouver, Canada. Association for Computational Linguistics.

\bibitem[{Artetxe et~al.(2018{\natexlab{a}})Artetxe, Labaka, and
  Agirre}]{artetxe2018robust}
Mikel Artetxe, Gorka Labaka, and Eneko Agirre. 2018{\natexlab{a}}.
\newblock \href {http://aclweb.org/anthology/P18-1073} {A robust self-learning
  method for fully unsupervised cross-lingual mappings of word embeddings}.
\newblock In \emph{Proceedings of the 56th Annual Meeting of the Association
  for Computational Linguistics (Volume 1: Long Papers)}, pages 789--798.
  Association for Computational Linguistics.

\bibitem[{Artetxe et~al.(2018{\natexlab{b}})Artetxe, Labaka, and
  Agirre}]{artetxe2018usmt}
Mikel Artetxe, Gorka Labaka, and Eneko Agirre. 2018{\natexlab{b}}.
\newblock \href {http://www.aclweb.org/anthology/D18-1399} {Unsupervised
  statistical machine translation}.
\newblock In \emph{Proceedings of the 2018 Conference on Empirical Methods in
  Natural Language Processing}, pages 3632--3642, Brussels, Belgium.
  Association for Computational Linguistics.

\bibitem[{Artetxe et~al.(2018{\natexlab{c}})Artetxe, Labaka, Agirre, and
  Cho}]{artetxe2018unmt}
Mikel Artetxe, Gorka Labaka, Eneko Agirre, and Kyunghyun Cho.
  2018{\natexlab{c}}.
\newblock \href {https://openreview.net/pdf?id=Sy2ogebAW} {Unsupervised neural
  machine translation}.
\newblock In \emph{Proceedings of the 6th International Conference on Learning
  Representations (ICLR 2018)}.

\bibitem[{Conneau et~al.(2018)Conneau, Lample, Ranzato, Denoyer, and
  J{\'{e}}gou}]{conneau2018word}
Alexis Conneau, Guillaume Lample, Marc'Aurelio Ranzato, Ludovic Denoyer, and
  Herv{\'{e}} J{\'{e}}gou. 2018.
\newblock \href {https://openreview.net/pdf?id=H196sainb} {Word translation
  without parallel data}.
\newblock In \emph{Proceedings of the 6th International Conference on Learning
  Representations (ICLR 2018)}.

\bibitem[{Dou and Knight(2012)}]{dou2012large}
Qing Dou and Kevin Knight. 2012.
\newblock \href {http://www.aclweb.org/anthology/D12-1025} {Large scale
  decipherment for out-of-domain machine translation}.
\newblock In \emph{Proceedings of the 2012 Joint Conference on Empirical
  Methods in Natural Language Processing and Computational Natural Language
  Learning}, pages 266--275, Jeju Island, Korea. Association for Computational
  Linguistics.

\bibitem[{Dou and Knight(2013)}]{dou2013dependency}
Qing Dou and Kevin Knight. 2013.
\newblock \href {http://www.aclweb.org/anthology/D13-1173} {Dependency-based
  decipherment for resource-limited machine translation}.
\newblock In \emph{Proceedings of the 2013 Conference on Empirical Methods in
  Natural Language Processing}, pages 1668--1676, Seattle, Washington, USA.
  Association for Computational Linguistics.

\bibitem[{Dou et~al.(2015)Dou, Vaswani, Knight, and Dyer}]{dou2015unifying}
Qing Dou, Ashish Vaswani, Kevin Knight, and Chris Dyer. 2015.
\newblock \href {http://www.aclweb.org/anthology/P15-1081} {Unifying bayesian
  inference and vector space models for improved decipherment}.
\newblock In \emph{Proceedings of the 53rd Annual Meeting of the Association
  for Computational Linguistics and the 7th International Joint Conference on
  Natural Language Processing (Volume 1: Long Papers)}, pages 836--845,
  Beijing, China. Association for Computational Linguistics.

\bibitem[{Dyer et~al.(2013)Dyer, Chahuneau, and Smith}]{dyer2013simple}
Chris Dyer, Victor Chahuneau, and Noah~A. Smith. 2013.
\newblock \href {http://www.aclweb.org/anthology/N13-1073} {A simple, fast, and
  effective reparameterization of ibm model 2}.
\newblock In \emph{Proceedings of the 2013 Conference of the North American
  Chapter of the Association for Computational Linguistics: Human Language
  Technologies}, pages 644--648, Atlanta, Georgia. Association for
  Computational Linguistics.

\bibitem[{Edunov et~al.(2018)Edunov, Ott, Auli, and
  Grangier}]{edunov2018understanding}
Sergey Edunov, Myle Ott, Michael Auli, and David Grangier. 2018.
\newblock \href {http://www.aclweb.org/anthology/D18-1045} {Understanding
  back-translation at scale}.
\newblock In \emph{Proceedings of the 2018 Conference on Empirical Methods in
  Natural Language Processing}, pages 489--500, Brussels, Belgium. Association
  for Computational Linguistics.

\bibitem[{Hassan et~al.(2018)Hassan, Aue, Chen, Chowdhary, Clark, Federmann,
  Huang, Junczys-Dowmunt, Lewis, Li et~al.}]{hassan2018achieving}
Hany Hassan, Anthony Aue, Chang Chen, Vishal Chowdhary, Jonathan Clark,
  Christian Federmann, Xuedong Huang, Marcin Junczys-Dowmunt, William Lewis,
  Mu~Li, et~al. 2018.
\newblock Achieving human parity on automatic chinese to english news
  translation.
\newblock \emph{arXiv preprint arXiv:1803.05567}.

\bibitem[{He et~al.(2016)He, Xia, Qin, Wang, Yu, Liu, and Ma}]{he2016dual}
Di~He, Yingce Xia, Tao Qin, Liwei Wang, Nenghai Yu, Tie-Yan Liu, and Wei-Ying
  Ma. 2016.
\newblock \href
  {http://papers.nips.cc/paper/6469-dual-learning-for-machine-translation.pdf}
  {Dual learning for machine translation}.
\newblock In \emph{Advances in Neural Information Processing Systems 29}, pages
  820--828.

\bibitem[{Heafield et~al.(2013)Heafield, Pouzyrevsky, Clark, and
  Koehn}]{heafield2013scalable}
Kenneth Heafield, Ivan Pouzyrevsky, Jonathan~H. Clark, and Philipp Koehn. 2013.
\newblock \href {http://www.aclweb.org/anthology/P13-2121} {Scalable modified
  kneser-ney language model estimation}.
\newblock In \emph{Proceedings of the 51st Annual Meeting of the Association
  for Computational Linguistics (Volume 2: Short Papers)}, pages 690--696,
  Sofia, Bulgaria. Association for Computational Linguistics.

\bibitem[{Lample and Conneau(2019)}]{lample2019cross}
Guillaume Lample and Alexis Conneau. 2019.
\newblock Cross-lingual language model pretraining.
\newblock \emph{arXiv preprint arXiv:1901.07291}.

\bibitem[{Lample et~al.(2018{\natexlab{a}})Lample, Conneau, Denoyer, and
  Ranzato}]{lample2018unsupervised}
Guillaume Lample, Alexis Conneau, Ludovic Denoyer, and Marc'Aurelio Ranzato.
  2018{\natexlab{a}}.
\newblock \href {https://openreview.net/pdf?id=rkYTTf-AZ} {Unsupervised machine
  translation using monolingual corpora only}.
\newblock In \emph{Proceedings of the 6th International Conference on Learning
  Representations (ICLR 2018)}.

\bibitem[{Lample et~al.(2018{\natexlab{b}})Lample, Ott, Conneau, Denoyer, and
  Ranzato}]{lample2018phrase}
Guillaume Lample, Myle Ott, Alexis Conneau, Ludovic Denoyer, and Marc'Aurelio
  Ranzato. 2018{\natexlab{b}}.
\newblock \href {http://www.aclweb.org/anthology/D18-1549} {Phrase-based \&
  neural unsupervised machine translation}.
\newblock In \emph{Proceedings of the 2018 Conference on Empirical Methods in
  Natural Language Processing}, pages 5039--5049, Brussels, Belgium.
  Association for Computational Linguistics.

\bibitem[{Levenshtein(1966)}]{levenshtein1966binary}
Vladimir~I Levenshtein. 1966.
\newblock Binary codes capable of correcting deletions, insertions, and
  reversals.
\newblock In \emph{Soviet physics doklady}, volume~10, pages 707--710.

\bibitem[{Marie and Fujita(2018)}]{marie2018unsupervised}
Benjamin Marie and Atsushi Fujita. 2018.
\newblock Unsupervised neural machine translation initialized by unsupervised
  statistical machine translation.
\newblock \emph{arXiv preprint arXiv:1810.12703}.

\bibitem[{McCallum et~al.(2005)McCallum, Bellare, and
  Pereira}]{mccallum2005conditional}
Andrew McCallum, Kedar Bellare, and Fernando Pereira. 2005.
\newblock A conditional random field for discriminatively-trained finite-state
  string edit distance.
\newblock In \emph{Proceedings of the Twenty-First Conference on Uncertainty in
  Artificial Intelligence}, pages 388--395.

\bibitem[{Mikolov et~al.(2013)Mikolov, Sutskever, Chen, Corrado, and
  Dean}]{mikolov2013distributed}
Tomas Mikolov, Ilya Sutskever, Kai Chen, Greg~S Corrado, and Jeff Dean. 2013.
\newblock \href
  {http://papers.nips.cc/paper/5021-distributed-representations-of-words-and-phrases-and-their-compositionality.pdf}
  {Distributed representations of words and phrases and their
  compositionality}.
\newblock In \emph{Advances in Neural Information Processing Systems 26}, pages
  3111--3119.

\bibitem[{Och(2003)}]{och2003MERT}
Franz~Josef Och. 2003.
\newblock \href {http://www.aclweb.org/anthology/P03-1021} {Minimum error rate
  training in statistical machine translation}.
\newblock In \emph{Proceedings of the 41st Annual Meeting of the Association
  for Computational Linguistics}, pages 160--167, Sapporo, Japan. Association
  for Computational Linguistics.

\bibitem[{Ott et~al.(2018)Ott, Edunov, Grangier, and Auli}]{ott2018scaling}
Myle Ott, Sergey Edunov, David Grangier, and Michael Auli. 2018.
\newblock \href {http://www.aclweb.org/anthology/W18-6301} {Scaling neural
  machine translation}.
\newblock In \emph{Proceedings of the Third Conference on Machine Translation:
  Research Papers}, pages 1--9, Belgium, Brussels. Association for
  Computational Linguistics.

\bibitem[{Post(2018)}]{post2018call}
Matt Post. 2018.
\newblock \href {http://www.aclweb.org/anthology/W18-6319} {A call for clarity
  in reporting bleu scores}.
\newblock In \emph{Proceedings of the Third Conference on Machine Translation:
  Research Papers}, pages 186--191, Belgium, Brussels. Association for
  Computational Linguistics.

\bibitem[{Ravi and Knight(2011)}]{ravi2011deciphering}
Sujith Ravi and Kevin Knight. 2011.
\newblock \href {http://www.aclweb.org/anthology/P11-1002} {Deciphering foreign
  language}.
\newblock In \emph{Proceedings of the 49th Annual Meeting of the Association
  for Computational Linguistics: Human Language Technologies}, pages 12--21,
  Portland, Oregon, USA. Association for Computational Linguistics.

\bibitem[{Ren et~al.(2019)Ren, Zhang, Liu, Zhou, and Ma}]{ren2019unsupervised}
Shuo Ren, Zhirui Zhang, Shujie Liu, Ming Zhou, and Shuai Ma. 2019.
\newblock Unsupervised neural machine translation with smt as posterior
  regularization.
\newblock \emph{arXiv preprint arXiv:1901.04112}.

\bibitem[{Sennrich et~al.(2016)Sennrich, Haddow, and
  Birch}]{sennrich2016improving}
Rico Sennrich, Barry Haddow, and Alexandra Birch. 2016.
\newblock \href {http://www.aclweb.org/anthology/P16-1009} {Improving neural
  machine translation models with monolingual data}.
\newblock In \emph{Proceedings of the 54th Annual Meeting of the Association
  for Computational Linguistics (Volume 1: Long Papers)}, pages 86--96, Berlin,
  Germany. Association for Computational Linguistics.

\bibitem[{Vaswani et~al.(2017)Vaswani, Shazeer, Parmar, Uszkoreit, Jones,
  Gomez, Kaiser, and Polosukhin}]{vaswani2017attention}
Ashish Vaswani, Noam Shazeer, Niki Parmar, Jakob Uszkoreit, Llion Jones,
  Aidan~N Gomez, {\L}ukasz Kaiser, and Illia Polosukhin. 2017.
\newblock Attention is all you need.
\newblock In \emph{Advances in Neural Information Processing Systems}, pages
  6000--6010.

\bibitem[{Yang et~al.(2018)Yang, Chen, Wang, and Xu}]{yang2018unsupervised}
Zhen Yang, Wei Chen, Feng Wang, and Bo~Xu. 2018.
\newblock \href {http://aclweb.org/anthology/P18-1005} {Unsupervised neural
  machine translation with weight sharing}.
\newblock In \emph{Proceedings of the 56th Annual Meeting of the Association
  for Computational Linguistics (Volume 1: Long Papers)}, pages 46--55.
  Association for Computational Linguistics.

\bibitem[{Zaidan(2009)}]{zaidan2009zmert}
Omar Zaidan. 2009.
\newblock Z-mert: A fully configurable open source tool for minimum error rate
  training of machine translation systems.
\newblock \emph{The Prague Bulletin of Mathematical Linguistics}, 91:79--88.

\bibitem[{Zhu et~al.(2017)Zhu, Park, Isola, and Efros}]{zhu2017unpaired}
Jun-Yan Zhu, Taesung Park, Phillip Isola, and Alexei~A. Efros. 2017.
\newblock Unpaired image-to-image translation using cycle-consistent
  adversarial networks.
\newblock In \emph{The IEEE International Conference on Computer Vision
  (ICCV)}.

\end{thebibliography}
\bibliographystyle{acl_natbib}

\end{document}